\documentclass{article}

\usepackage[preprint]{neurips_2023}
\usepackage{graphicx}
\usepackage{fontawesome}

\usepackage[utf8]{inputenc} 
\usepackage[T1]{fontenc}    
\usepackage{epigraph}
\usepackage{url}            
\usepackage{booktabs}       
\usepackage{amsfonts}       
\usepackage{listings}
\usepackage{xcolor}         

\usepackage{nicefrac}

\usepackage{microtype}      
\usepackage{xspace}
\usepackage{multirow}
\usepackage{pifont}
\newcommand{\baby}{\textsc{Agents}\xspace}

\usepackage{appendix}
\newcommand{\yes}{\color{green!60!black}\ding{51}} 
\newcommand{\no}{\color{red!60!black}\ding{55}}   

\title{Agents: An Open-source Framework \\ for Autonomous Language Agents}

\author{ 
\textbf{Wangchunshu Zhou}\textsuperscript{1}\thanks{Equal Contribution. Correspondence to: chunshu@aiwaves.cn}\quad \textbf{Yuchen Eleanor Jiang}\textsuperscript{1*} \quad \textbf{Long Li}\textsuperscript{1*}\quad \textbf{Jialong Wu}\textsuperscript{1*}\\
\textbf{Tiannan Wang}\textsuperscript{1}\quad 
\textbf{Shi Qiu}\textsuperscript{1}\quad \textbf{Jintian Zhang}\textsuperscript{1}\quad \textbf{Jing Chen}\textsuperscript{1}\quad \textbf{Ruipu Wu}\textsuperscript{1}\quad 
\textbf{Shuai Wang}\textsuperscript{1}\quad \\ 
\textbf{Shiding Zhu}\textsuperscript{1}\quad 
\textbf{Jiyu Chen}\textsuperscript{1}\quad \textbf{Wentao Zhang}\textsuperscript{1}\quad
\textbf{Xiangru Tang}\quad
\textbf{Ningyu Zhang}\textsuperscript{2}\quad \textbf{Huajun Chen}\textsuperscript{2}\quad\\
\textbf{Peng Cui}\textsuperscript{3}\quad \textbf{Mrinmaya Sachan}\textsuperscript{3}
\\
\\
\textsuperscript{1}\textbf{AIWaves Inc.} \quad \textsuperscript{2}\textbf{Zhejiang University}
\quad \textsuperscript{3}\textbf{ETH Z\"{u}rich}
}

\begin{document}

\maketitle

\begin{abstract}
Recent advances on large language models (LLMs) enable researchers and developers to build autonomous language agents that can automatically solve various tasks and interact with environments, humans, and other agents using natural language interfaces. We consider language agents as a promising direction towards artificial general intelligence and release \baby, an open-source library with the goal of opening up these advances to a
wider non-specialist audience. \baby is carefully engineered to support important features including \textit{planning}, \textit{memory}, \textit{tool usage}, \textit{multi-agent communication}, and \textit{fine-grained symbolic control}. \baby is user-friendly as it enables non-specialists to build, customize, test, tune, and deploy state-of-the-art autonomous language agents without much coding. 
The library is also research-friendly as its modularized design makes it easily extensible for researchers. \baby is available at \url{https://github.com/aiwaves-cn/agents}.
\end{abstract}
\section{Introduction}

\vspace{-3pt}
\setlength{\epigraphwidth}{0.95\columnwidth}
\epigraph{``An autonomous agent is a system situated within and a part of an environment that senses the environment and acts on it, over time, in pursuit of its own agenda and so as to effect what it senses in the future.''}
{\textit{Is it an Agent, or just a Program?: A Taxonomy for Autonomous Agents~\citep{franklin1996agent}}}

\begin{figure}[t]
    \centering
   \includegraphics[width=\textwidth,trim={0cm 0cm 0cm 0cm} ,clip]{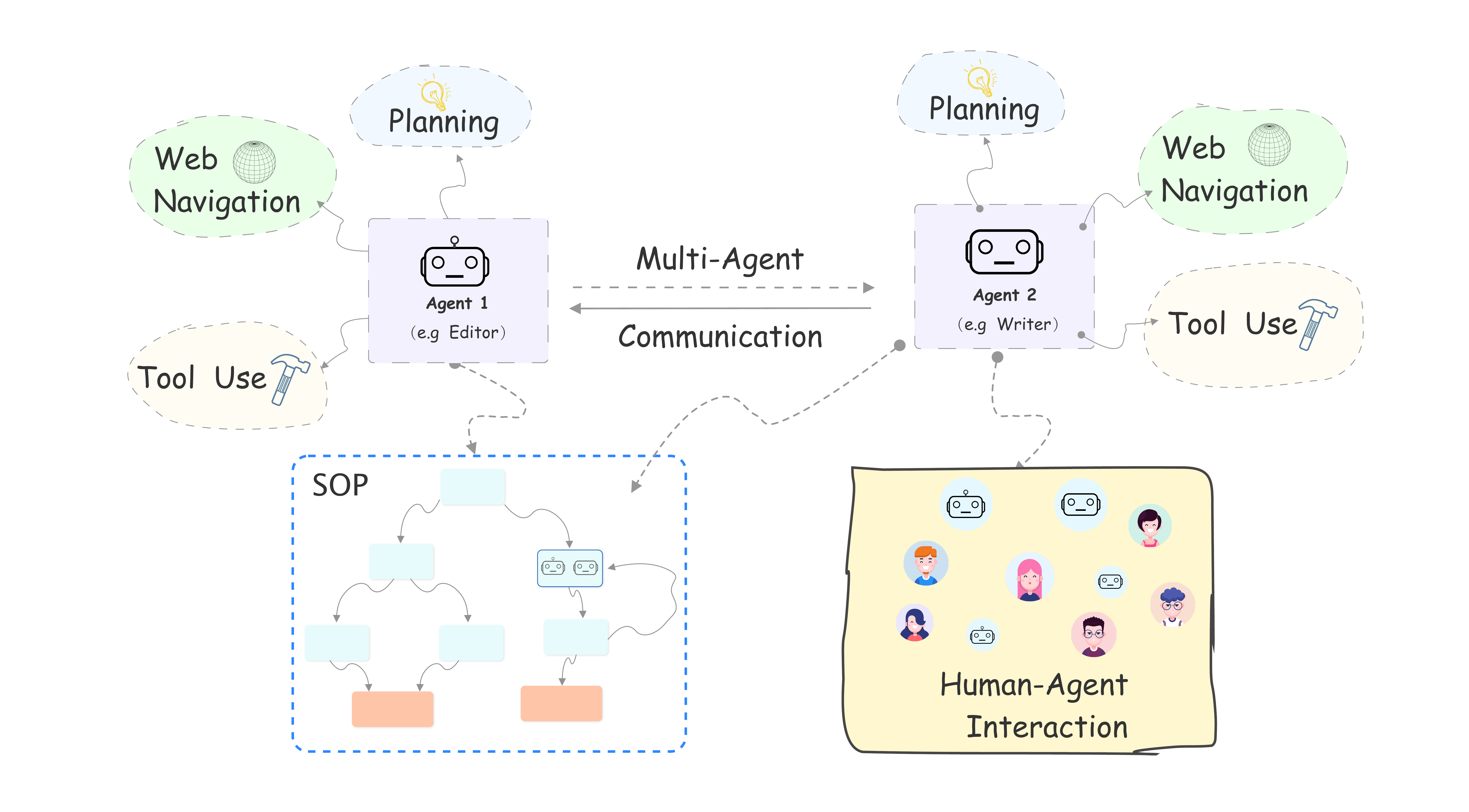}
    \caption{Illustration of the \baby framework. \looseness=-1}
    \label{fig:main}
\end{figure}

Large Language Models (LLMs)~\citep{gpt3, gpt3.5, gpt4} such as ChatGPT make it possible to build autonomous agents that can automatically solve complicated tasks and interact with the environment, humans, or other agents by perceiving, reasoning, planning, and acting in the world~\citep{weng2023prompt}. Language agents are a promising step towards artificial general intelligence (AGI) and can help reduce human effort in certain roles such as customer service, consulting, programming, writing, teaching, etc.  Some recent demos such as AutoGPT~\citep{autogpt} and BabyAGI~\citep{babyagi} have demonstrated the potential of language agents and have gained massive interest from developers, researchers, as well as more non-technical audiences. 

While intriguing, most of these demos or repositories are not friendly for \textit{customizing}, \textit{tuning}, and \textit{deploying} new agents even for experienced developers or researchers. This limitation comes from the fact that these demos typically proof-of-concepts showcasing the possibility of language agents, instead of being larger frameworks that can be used to build and customize language agents over time. Moreover, most of these open-source repositories only cover a small portion of the core abilities of language agents including task decomposition~\citep{nye2022show}, long-short term memory~\citep{zhou2023recurrentgpt}, web navigation~\citep{webgpt}, tool usage~\citep{toolformer}, and multi-agent communication~\citep{multi-agent-rl}. In addition, most (if not all) existing language agent frameworks solely depend on a short task description and rely completely on the abilities of LLMs to plan and act. This results in significant randomness and inconsistency across different runs, delivering an unsatisfactory user experience and making it hard to customize and tune language agents.

We believe the aforementioned limitations are important barriers for recent advances in language agents to reach a broader non-specialist audience and impact our society in a positive way. To this end, we release \baby, an open-source library and framework for language agents dedicated to supporting LLM-powered language agents. \baby's philosophy is to make customizing, tuning, and deploying language agents as simple as possible even for non-specialists while also remaining easily extensible for developers and researchers. In addition, the library also provides the following key features that make it a versatile framework for language agents:
\paragraph{Long-short term memory}
According to~\citet{franklin1996agent}, a key difference between autonomous agents and computer programs (or machine learning models) is that machine learning models only need to respond to a single input/query, while autonomous agents need to interact with environments or other agents over time. Therefore, the ability to maintain long-short term memory is very important for autonomous agents. \baby integrates the memory components in~\citep{zhou2023recurrentgpt} and enables language agents to store and retrieve long-term memory with VectorDB and semantic search, and regularly update a short-term working memory with a scratchpad. Users can choose to equip an agent with long-term memory, short-term memory, or both of them by simply filling in a field in the config file.
\paragraph{Tool usage \& Web navigation}
Another important feature for autonomous agents is the ability to use external tools and surf the internet. This is especially important for language agents because they rely on the language interface and thus need to use external tools to interact with environments beyond language communication and navigate the web to gather useful information. Following~\citep{patil2023gorilla}, \baby supports a few commonly used external APIs and provides an abstract class that enables developers to integrate other tools with ease. We also enable agents to navigate the internet and gather information by defining web search and web navigation as specialized APIs.

\paragraph{Multi-agent communication}
In addition to single-agent abilities, \baby also supports customizing multi-agent systems, which can be helpful for certain applications such as games~\citep{park2023generative}, social experiments~\citep{li2023camel}, software development~\citep{qian2023communicative}, etc. One new feature for multi-agent communication in \baby is the ``dynamic scheduling'' feature. Instead of scheduling the order for the agents to act with hard-coded rules, dynamic scheduling provides an option to define a controller agent that acts as a ``moderator'' and decides which agent to perform the next action considering their roles and the current history. Dynamic scheduling has the potential to make communication between multiple agents more natural and flexible. Developers can easily customize the controller by specifying its rule in the config file using natural language.

\paragraph{Human-agent interaction}
One limitation in existing agent frameworks is that while they enable agents, or multi-agents, to automatically solve tasks, it's not easy or even possible for human users to interact with the agents, especially in the multi-agent scenario. \baby seamlessly supports human-agent interaction in both single-agent and multi-agent scenarios, making it possible for one or more humans to communicate and interact with language agents.

\paragraph{Controllabilty}
Existing agent frameworks generally define and control the agents' behavior only using a system prompt and then let the agent plan and act on its own. In contrast, \baby provides a novel paradigm to build \textit{controllable agents} via a \textit{symbolic plan}, also referred to as \textit{standard operating procedures} (SOPs). An SOP is a graph of multiple states that defines different situations an agent may encounter while accomplishing a task, and the transition rules between the states. Similar to SOPs in the real world, an SOP in \baby is a meticulously documented set of step-by-step instructions that outlines how a particular task or process should be performed by an agent or a group of agents. SOPs can be generated by an LLM and edited by the user when customizing and tuning the agent. After deployment, an agent will behave following specified instructions and guidelines for each state and dynamically adjust its current state according to its interaction with the environment, humans, or other agents. The introduction of the symbolic plan offers the opportunity to provide fine-grained control of an agent's behavior, making agents' behavior more stable/predictable and facilitating tuning/optimizing agents at the same time.

In addition, we propose an automated SOP generation pipeline to reduce human labor on writing detailed SOP and config files when customizing (multi-) agent systems. The automated SOP generation pipeline is a ``meta agent'' that can generate config files for language agents with retrieval-augmented generation given a short description of the task.

\baby is an ongoing effort maintained by researchers and engineers from AIWaves\footnote{\url{https://www.aiwaves.org/}}. We look forward to support from community contributors on the project. The library and detailed documentation and tutorials are available on GitHub\footnote{\url{https://github.com/aiwaves-cn/agents}}.

\section{Related Work}
\subsection{Autonomous Language Agents}
The concept of language agents has become very popular recently and a variety of language agents targeting different tasks have been proposed. For example, Generative Agents~\citep{park2023generative} developed language agents to mimic human social behavior, WebAgent~\citep{gur2023realworld} demonstrated the possibility to build language agents that can complete the tasks on real websites following natural language instructions,~\citet{qian2023communicative} and MetaGPT~\citep{hong2023metagpt} experimented with software development in multi-agent communication settings, and ~\citet{zhou2023recurrentgpt} built language agents that act as interactive writing assistants.

In addition to language agents that target specific tasks, recent open-source projects such as AutoGPT~\citep{autogpt}, BabyAGI~\citep{babyagi}, and SuperAGI~\citep{superagi} are aimed at the goal of building autonomous agents that do whatever users want and attracted massive interest from both developers and non-specialist audiences.

\begin{table}[htbp]
\centering
\caption{Comparison of Language Agent Frameworks}
\label{tab:comparison}
\resizebox{\textwidth}{!}{
\begin{tabular}{l c c c c c c c}
\toprule
\textbf{Framework} & \textbf{Tool Usage} & \textbf{Long-short Term Memory} & \textbf{Multi-Agent} & \textbf{Human-Agent Interaction} & \textbf{Symbolic Control} \\
\midrule
Transformers Agents &  \yes & \no & \no & \no & \no \\
LangChain &  \yes & \yes & \no & \no & \no  \\
Auto-GPT &  \yes & \no & \no & \no & \no  \\
Gentopia &  \yes & \no & \no & \no & \no  \\
XLang &  \yes & \no & \no & \no & \no  \\
Meta-GPT &  \yes & \no & \yes & \no & \no  \\
Camel &  \yes & \no & \yes & \no & \no  \\
AgentVerse &  \yes & \yes & \yes & \yes & \no  \\
\midrule
\baby &  \yes & \yes & \yes & \yes & \yes  \\
\bottomrule
\end{tabular}}

\end{table}

\subsection{Language Agents Frameworks}
More recently, a few open-source frameworks for language agents have been proposed. For example, Transformers Agents~\citep{transformers} builds language agents that can automatically use tools to solve tasks described in natural language; LangChain~\citep{langchain} supports end-to-end language agents that can automatically solve tasks specified in natural language; Camel~\citep{li2023camel} and AgentVerse~\citep{chen2023agentverse} are platforms tailored for building multi-agent systems; Gentopia~\citep{xu2023gentopia} and XLang\footnote{\url{https://github.com/xlang-ai/xlang}} are libraries for building tool-augmented agents. We illustrate the key features supported by these platforms and \baby in Table 1. We can see that \baby is the only framework that supports tool usage, long-short term memory, and multi-agent communication at the same time. \baby also offers human-agent interaction and controllability through symbolic plans (SOPs) for the first time.

\section{Library Design}

\begin{lstlisting}[caption=Exemplar code for initializing and running a (multi) agent system with \baby]
def main()
    # agents is a dict of one or multiple agents.
    agents = Agent.from_config("./config.json")
    sop = SOP.from_config("./config.json")
    environment = Environment.from_config("./config.json")
    run(agents,sop,environment)
\end{lstlisting}

\baby is designed following the philosophy in~\citet{franklin1996agent}: \textit{``an \textbf{autonomous agent} is situated in an \textbf{environment}''}. Therefore, \textbf{agent} and \textbf{environment} are two major classes in the \baby framework. In addition to these two classes, we also include a class for symbolic plans, named \textbf{SOP} (short for Standard Operating Procedure), to make language agents more controllable. These main classes are all initialized from a config file which can be filled in plain text. In sum, a typical script for initializing and running a (multi) agent system with \baby is illustrated in Code 1. The config file not only defines these core objects but also factorizes complicated prompts into modularized prompt components. The factorization of prompts significantly reduces the expertise requirements and efforts for users to build (multi) agent systems. Using a single config file to define the agents, plan, and basic environment also facilitates the sharing of language agents (which will be discussed in the Agent Hub section). Each of these three core classes consist of standardized APIs that can be overwritten by experienced developers and researchers. We describe these classes in detail:

\begin{lstlisting}[caption=Exemplar code for the running loop of a (multi) agent system in \baby]
def run(agents,sop,environment):
    while not sop.finished:
        agent,state=sop.step(agents, environment)
        action=agent.step(state,environment)
        environment.update(agent,action)
        # optional, in case of dynamic planning
        # new_states = get_new_states(action)
        # sop.add_states(new_states)
\end{lstlisting}

\subsection{Agent}
The \textit{Agent} class abstracts a language agent. Its UML is illustrated in Figure 1. We can see that an agent maintains its long-short term memory and has methods to observe the environment (\texttt{agent.\_observe(environment)}), act according to its current state (\texttt{agent.\_act()}) and update its memory (\texttt{agent.\_update\_memory()}). All these methods are wrapped in the \texttt{agent.step()} method. This factorization enables developers to customize agents with new functionalities easily. Unlike existing language agent frameworks that assume an agent must be based on an LLM, we include a ``\texttt{\_is\_human}'' property to an agent. If it is set to ``\texttt{True}'', the (\texttt{agent.\_act()}) will opt to provide observations and memory information to a human user and wait for the human user to input the action. This design allows flexible human-agent interaction in both single-agent and multi-agent systems by allowing human users to take the role of one or more language agents. It facilitates developers to build various interesting applications such as allowing human users to act as members of a team in debate and collaborate with (agent or human-based) teammates to beat another team, or act as CTO/engineers in a software company and collaborate with others for software development.

\subsection{SOP}
The SOP class contains a graph of the states of agents. Each state specifies a certain sub-task or sub-goal of all agents when accomplishing the task described by the SOP. States are abstracted into a \texttt{State} class. A \texttt{State} object contains modularized prompts for the agent to leverage an LLM and various tools or APIs that an agent can use in the state. We abstract everything an agent may use for action in a state into a ``\texttt{Component}'' class. The \texttt{Component} class consists of two subclasses corresponding to different parts of the prompt and tools or external APIs, named ``\texttt{PromptComponent}'' and ``\texttt{ToolComponent}'', respectively. \texttt{PromptComponent} includes modularized prompts that specify the task/goal, rules/constraints, (step-by-step) demonstrations for in-context learning, and the output format. \texttt{ToolComponent} supports more complex usage beyond modularized prompts, including external tools and APIs such as web search, knowledge bases, etc. The results of the tools are either included in the prompt or directly returned and processed afterward, according to the config file.

An SOP object also includes an LLM-based control function that decides the transition between different states and the next agent to act. The state transit function is named \texttt{sop.\_transit()} and the agent routing function is named \texttt{sop.\_route()}. Both of the functions are wrapped in an \texttt{sop.next()} function which is used in the main loop.

\begin{figure}[htbp]
	\centering
	\begin{minipage}{0.49\linewidth}
		\centering
		\includegraphics[width=0.9\linewidth]{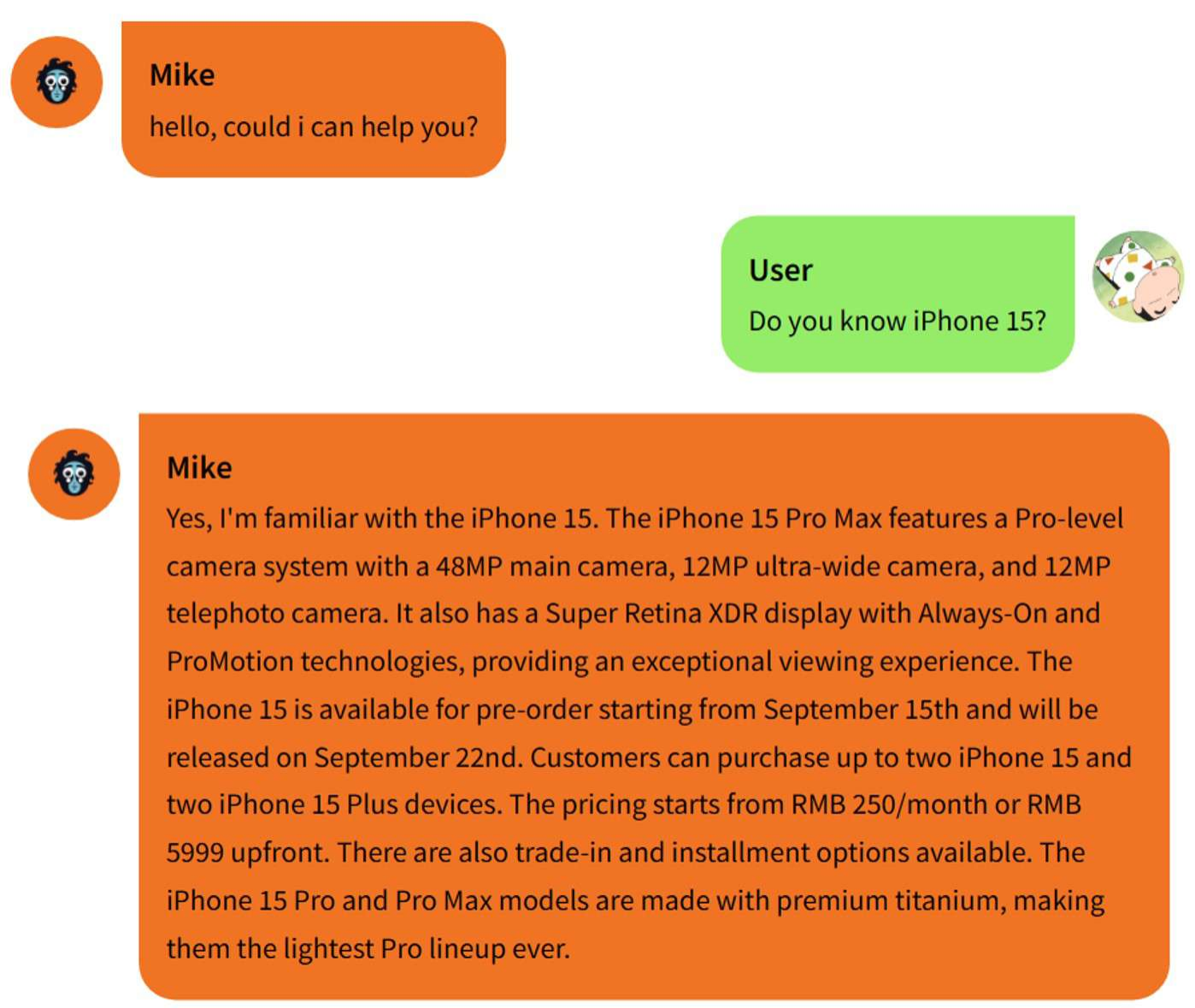}
            \caption{(a) Customer service agent}
            \label{customer-service}
	\end{minipage}
	\begin{minipage}{0.49\linewidth}
		\centering
		\includegraphics[width=0.965\linewidth]{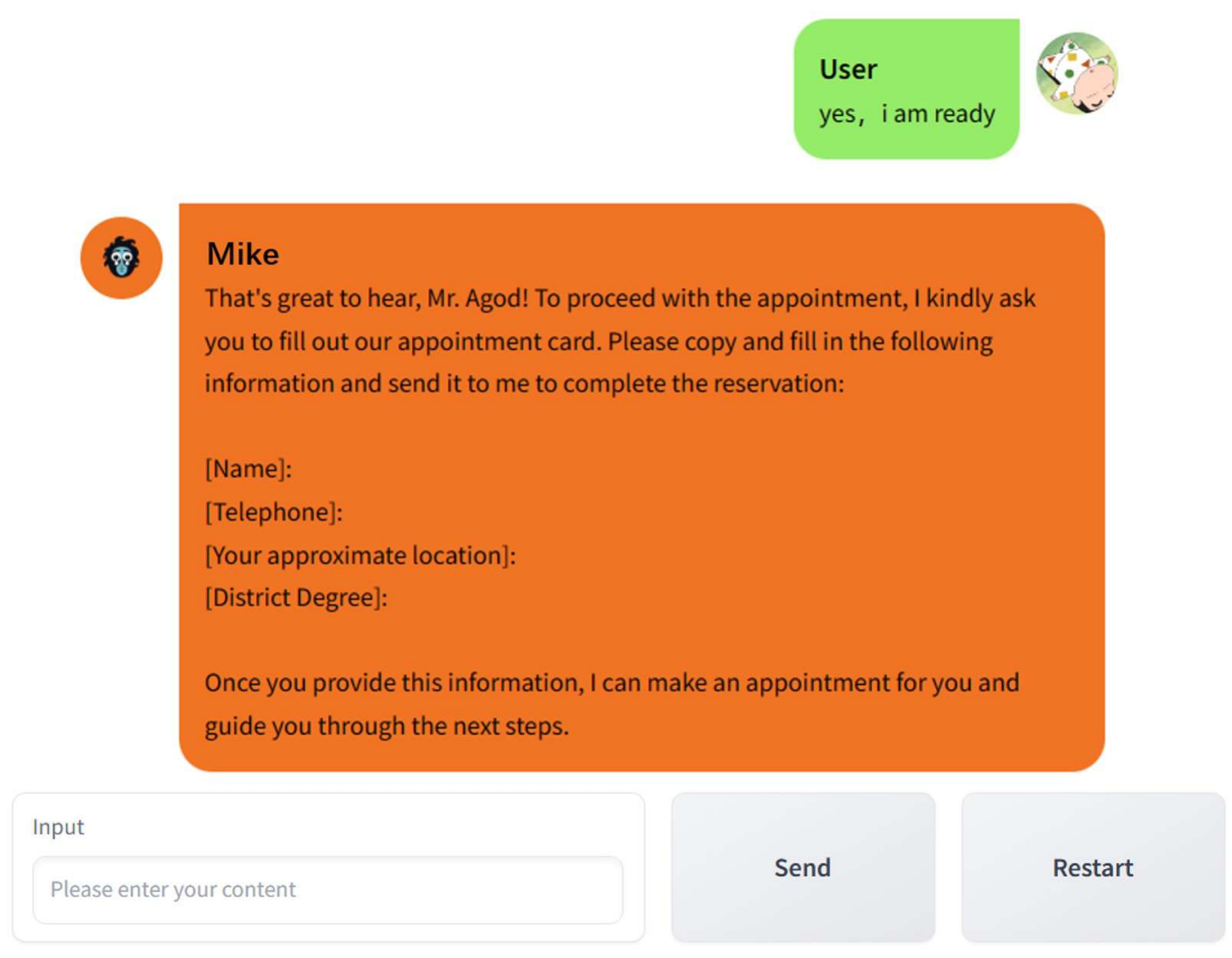}
              \caption{(b) Sales agent}
              \label{sales}
	\end{minipage}
\end{figure}

\subsection{Environment}

The \texttt{Environment} class abstracts the environment in which the agents are situated. An \texttt{environment} consists of two main functions: \texttt{environment.\_observed()} and \texttt{environment.update()}. \texttt{environment.\_observed()} defines how the environment influences the agent's action (i.e., what information should be transferred to the agent upon observation, and \texttt{environment.update()} defines how the agent's action impacts the environment.

The execution logic of a (multi) agent system based on \baby is very intuitive. As illustrated in Code 2, in each iteration, the SOP first decides the state transition and selects the next agent to act based on the agents and the environment. The agent then takes an action based on its state and the environment. Then the environment updates itself based on the new action. Finally, if a workflow requires dynamically adjusting the plan based on the intermediate execution results, one can parse the output from an action, define a new state and add it into the current SOP.

\subsection{Implementation Details of Core Features}

\paragraph{Long-short Term Memory}: \baby implements long-short term memories for language agents following~\citet{zhou2023recurrentgpt}. Specifically, long-term memories are action histories and are embedded by sentence-transformers~\citep{reimers-2019-sentence-bert}, stored in a VectorDB, and queried via semantic search. Short-term memories, or working memories, are in natural language form and updated by an LLM via a carefully tuned prompt.

\paragraph{Tool Usage \& Web Navigation}: \baby supports tool usage and web navigation via \texttt{ToolComponents}. For each external tool or API, developer can wrap the API call in the\texttt{ToolComponent.func()} method. For complicated tools of which the API call is context-dependent, \baby integrates the 
the ``Function-calling'' feature of OpenAI's GPT APIs to let LLMs decide how to use the tools. Web navigation is achieved by implementing web search as a specialized tool.

\begin{figure}[t]
    \centering
   \includegraphics[width=\textwidth,trim={0cm 0cm 0cm 0cm} ,clip]{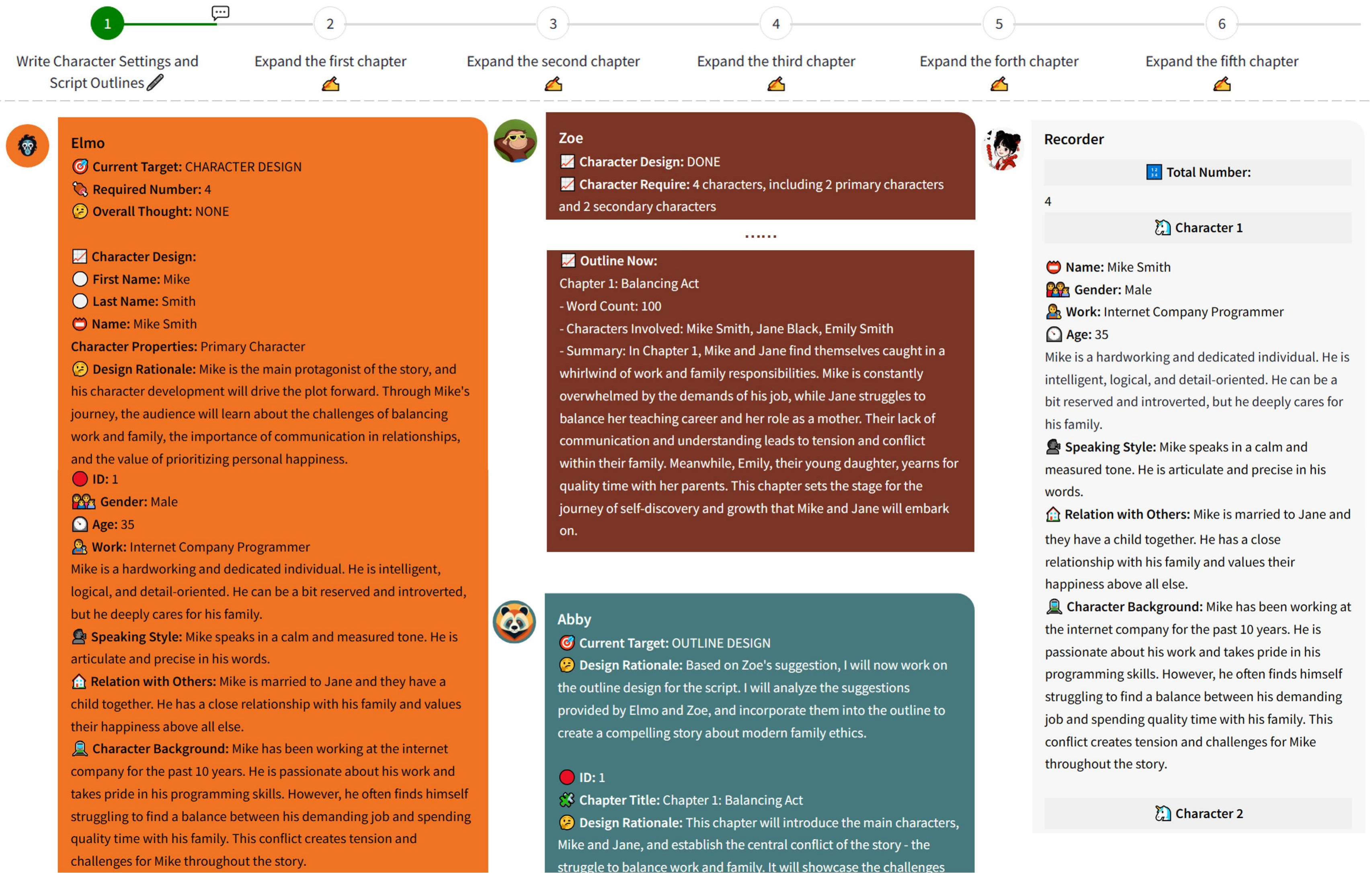}
    \caption{Multi-Agent System: Fiction Studio. \looseness=-1}
    \label{fig:multi-agent}
\end{figure}

\paragraph{Multi-Agent Communication}: Different from most existing frameworks for multi-agent systems that use pre-defined rules (e.g., let each agent act in a sequential order) to control the order for agents' action, \baby includes a controller function that dynamically decides which agent will perform the next action using an LLM by considering the previous actions, the environment, and the target of the current states. This makes multi-agent communication more flexible.

\paragraph{Human-Agent Interaction}: \baby supports human-agent interaction in multi-agent systems by allowing human users to change the ``\texttt{is\_human}'' field for a certain agent in the config file to ``\texttt{True}''. In this case, the user can play the role of the agent by himself/herself and input his/her own actions and interact with other language agents in the environment.

\subsection{Deployment}

Existing open-source frameworks for language agents focus on building proof-of-concept language agents that run either in the terminal or on Gradio~\citep{abid2019gradio}. In contrast, \baby supports deploying language agents as APIs with FastAPI\footnote{https://fastapi.tiangolo.com/}. This greatly facilitates developers to integrate language agents in real-world applications.

\subsection{The Agent Hub}

\baby aims to not only facilitate the development, testing, and tuning of a language agents system but also makes the distribution and sharing of language agents easier. To this end, we introduce \textsc{Agent Hub}, a platform that allows users to share their fine-tuned language agents as well as search/download useful language agents that others share on the platform. In this way, one can easily customize language agents by starting from community agents and slightly modifying them. This greatly reduces the effort of designing, testing, and tuning language agents from scratch. 

\subsection{Automatic Creation of Agent Systems}
While using an SOP to provide fine-grained control to language agents, it can sometimes be laborsome for users to manually specify the SOP from scratch since it requires to set different states, their connections, and the prompts and tools for each \texttt{Component} for all states. Therefore, we carefully implement a pipeline for automatic SOP generation. Our SOP generation framework is based on retrieval-augmented generation (RAG)~\citep{DBLP:conf/nips/LewisPPPKGKLYR020}. The SOP generation pipeline itself is also based on the \baby framework and has an SOP of first specifying the agents required, then planning the states and their connections, and finally generating the \texttt{Components}. Therefore, this pipeline can be regarded as a ``meta agent'' that can create other agents and multi-agent systems. Detailed description of the automatic agent creation framework is decribed in~\citep{zhou2023towards}.
\begin{figure}[t]
    \centering
   \includegraphics[width=\textwidth,trim={0cm 0cm 0cm 0cm} ,clip]{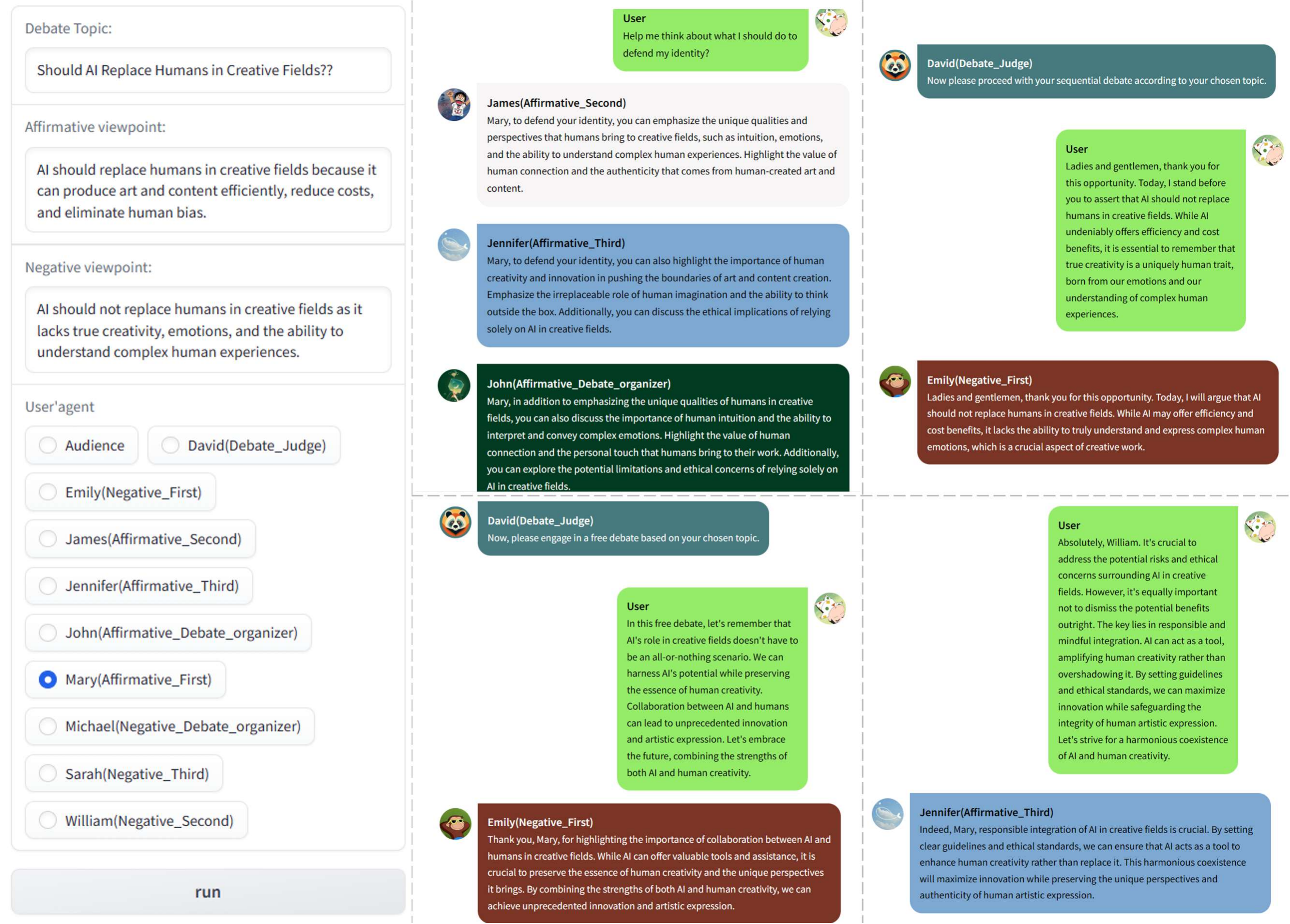}
    \caption{Human-Agent Interaction in a debate. \looseness=-1}
    \label{fig:human-agent}
\end{figure}


\section{Case Studies}

We then present a few case studies on different language agents built with the library, including single-agent systems, multi-agent systems, and systems that require human-agent interaction. All demos are available at \url{http://www.aiwaves-agents.com/}.

\subsection{Single-agent systems}
We implement a few single-agent systems with \baby including a chit-chat bot, two customer service agents based on knowledge bases and web search engines, a shopping assistant agent, and a sales agent. The agents demonstrate different features of the library and the possibility of building language agents of different use cases using \baby. We present a screenshot of the customer service agent and the sales agent in Figure \ref{customer-service} and \ref{sales}, respectively.

\subsection{Multi-agent systems}
We also demonstrate how one can build a multi-agent system consisting of multiple agents interacting with each other in an environment. We select three scenarios including a fiction studio, a debate, and a software company. These scenarios include both cooperative and competitive scenarios, which are two main categories of multi-agent systems. All of the scenarios include multiple subtasks that are controlled through symbolic plans, i.e., SOPs. One can easily observe the language agents' behavior in each subtask and engineer the corresponding prompts to customize and improve the system. We present a system screenshot of the fiction studio system in Figure \ref{fig:multi-agent}. We also showcase the human-agent interaction feature of the framework in a case study where a human user participate in a debate with language agents in Figure \ref{fig:human-agent}.
\section{Conclusion}

LLMs and language agents powered by them are playing increasingly important roles in both the NLP/AI community and our society in general. \baby, is a unified framework and open-source library for language agents. \baby aims to facilitate developers to build applications with language agents, researchers to conduct language agents research, and general non-technical audiences to build and customize personalized language agents.

\bibliographystyle{unsrtnat}
\bibliography{reference}

\newpage

\appendix



\end{document}